\documentclass[runningheads]{llncs}
\usepackage[T1]{fontenc}
\usepackage{graphicx}
\usepackage{color}
\usepackage{hyperref}

\begin{document}

\title{RadYOLO: Computationally Efficient 3D Object Detection and Segmentation in CT and MRI}

\titlerunning{RadYOLO: Efficient 3D object detection}

\author{Kai Geissler\orcidID{0009-0008-9709-0335} \and
Laurens Müller-Groh\orcidID{0009-0000-1934-5365} \and
Hans Meine\orcidID{0000-0002-7557-5007}}

\authorrunning{K. Geissler et al.}

\institute{Fraunhofer Institute for Digital Medicine MEVIS, Max-von-Laue-Str. 2, 28359 Bremen, Germany \\
\email{kai.geissler@mevis.fraunhofer.de}}

\maketitle

\begin{abstract}
Object detection and segmentation in three-dimensional medical images is a very active area of research. 
However, most proposed deep learning models carry a high computational cost, and only few aim to be broadly applicable, achieve high detection performance, and remain fast to execute on resource-constrained hardware.
To address this gap, we present RadYOLO, a 3D extension of YOLO11 tailored to medical images.
We compare it with nnU-Net and nnDetection on five datasets comprising CT and MRI data with varying object sizes and prevalence.
RadYOLO's detection performance surpasses that of nnDetection on four of five datasets and is comparable on one.
Compared to nnU-Net, RadYOLO performs better on lesion detection tasks, while nnU-Net excels at detecting large organs when precise localization is required.
When rough object localization is sufficient, RadYOLO matches or outperforms nnU-Net on all five datasets.
Regarding inference time, RadYOLO is 8–46× faster than nnU-Net on a GPU.
Compared to nnDetection the speedup is even higher.
When executed on a CPU, RadYOLO's inference runs within seconds (still faster than nnU-Net on a GPU) offering a significant advantage for clinical and edge-device deployment.

RadYOLO repository: \href{https://github.com/FraunhoferMEVIS/RadYOLO}{github.com/FraunhoferMEVIS/RadYOLO} 

\keywords{Efficient medical object detection \and Computationally efficient medical image analysis \and Efficient segmentation medical}
\end{abstract}

\section{Introduction}
Numerous architectures for object detection and segmentation in 3D medical images have been proposed, with new ones appearing continuously.
However, most focus purely on detection or segmentation accuracy and either neglect computational efficiency entirely or treat it as a side note.
For real-world deployment, however, computational cost is a critically important model property.
It directly determines the serving cost, prediction latency, and the range of hardware on which a model can run within acceptable time.

In the 2D image analysis field, the YOLO model family dominates efficient object detection. 
After the original YOLO~\cite{redmon2016you}, many modifications and improved versions have been proposed, with YOLO11~\cite{yolo11_ultralytics} and YOLO26~\cite{jocher2026ultralytics} being two of the most recent ones.

The literature on efficient object detection for 3D medical imaging is comparatively sparse.
MedYOLO~\cite{sobek2024medyolo} extended YOLOv5~\cite{yolov5} to 3D medical images but exhibited limited detection performance on some datasets, an issue subsequently investigated and addressed by RevisedMedYOLO~\cite{geissler2025revisedmedyolo}.

Other approaches use pseudo 3D models (or 2+1 models) for efficient object detection~\cite{cai2020deep,zhang2020revisiting} that can partially leverage models from the 2D image processing field. 
A recent Transformer-based model for object detection in radiological volumetric images, Organ-DETR~\cite{ghahremani2025organ}, can process several large 3D patches per second on a strong GPU, but has not been evaluated on CPU.
In the field of biomedical imaging, Mask R-CNN was extended to 3D Mask R-CNN~\cite{david2025end}.
However its inference time was reported to be 13 hours for 100 samples using an Nvidia V100 GPU.
In the field of medical image segmentation, efficient architectures are also investigated. 
For example UNETR++~\cite{shaker2024unetr++} can process patches of size $128^3$ in 1.5 seconds on a CPU.
However, as it operates on restricted patches, the total inference time for a whole imaging volume can quickly grow large.

In this work we present RadYOLO, an extension of the YOLO11 model to 3D medical images.
RadYOLO is fast on both GPU and CPU, accurate, and provides joint detection, classification, and segmentation of diverse anatomical structures in medical images.
We evaluate it on five different datasets to showcase its broad applicability.

\section{Methods}
\subsection{Data}
We use five public datasets to evaluate RadYOLO and compare it against other deep learning architectures. 
These datasets were selected to represent different imaging modalities, image sizes, and types of anatomical targets.
AMOS22~\cite{ji2022amos} comprises 300 CT and 60 MRI scans (which are publicly available) with segmentation masks for 15 abdominal organs. 
Liver Lesions~\cite{nicoli2025liver} is a dataset of 842 CT scans with segmentation masks for liver lesions. LUNA16~\cite{setio2017validation} consists of 888 CTs with lung nodule masks.
MAMA-MIA~\cite{garrucho2025large} is a dynamic contrast-enhanced breast MRI dataset with annotations of the primary tumor in each volume.
The VerSe~\cite{sekuboyina2021verse} dataset offers vertebra segmentations for 355 CT scans.
\tablename~\ref{tab:datasets} provides an overview of these datasets. 
All datasets were split 60/20/20\% into training, validation, and test sets for model training, model selection, and final evaluation, respectively.

\begin{table}[tb]
\caption{Dataset overview. Objects shows the average number of objects per image.}
\label{tab:datasets}
\centering
\begin{tabular}{|l|c|c|l|r|r|}
\hline
Name & Modality & Train/Val/Test Cases & Target structures & Classes & Objects \\
\hline
AMOS22~\cite{ji2022amos} & CT\&MR & 216 / 72 / 72 & Abdominal organs & 15 & 14.5 \\
Liver Lesions~\cite{nicoli2025liver} &  CT & 498 / 167 / 167 & Liver lesions & 1 & 9.0 \\
LUNA16~\cite{setio2017validation} & CT & 532 / 178 / 178 & Lung nodules & 1 & 1.8 \\
MAMA-MIA~\cite{garrucho2025large} & MR & 903 / 301 / 302 & Breast lesions & 1 & 1.0 \\
VerSe~\cite{sekuboyina2021verse} & CT & 213 / 71 / 71 & Vertebrae & 26 & 12.3 \\
\hline
\end{tabular}
\end{table}

\subsection{RadYOLO}
\begin{figure}[tb]
\centering
\includegraphics[width=\textwidth,page=1]{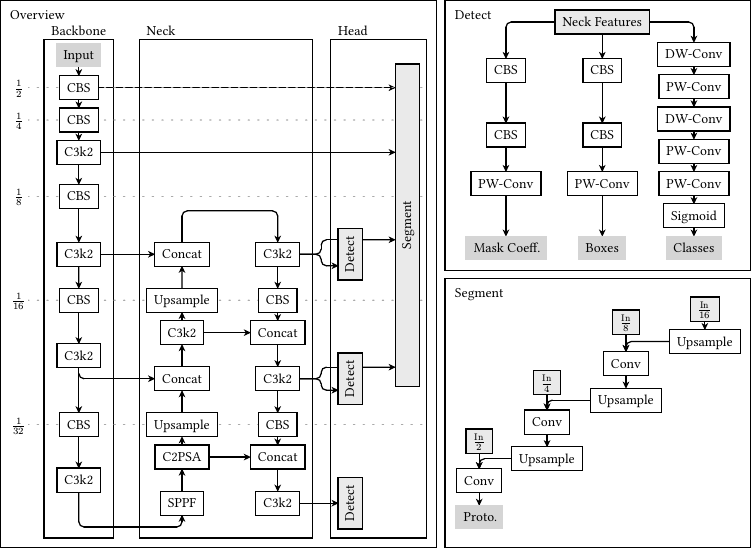}
\caption{RadYOLO architecture overview. It consists of a backbone for feature extraction, a neck to create a feature pyramid and detect and segment blocks for final predictions. CBS: Convolution, Batch normalization, SiLU activation. C3k2: Special combination of CBS layers~\cite{jocher2026ultralytics}. SPPF: Spatial Pyramid Pooling Fast layer. C2PSA: Convolution and Partial Spatial Attention. PW-Conv: Point-Wise Convolution. DW-Conv: Depth-Wise Convolution. Mask Coeff.: Mask Coefficients. Proto.: Prototype Masks.}
\label{fig:radyolo_architecture_overview}
\end{figure}

RadYOLO is an extension of YOLO11~\cite{yolo11_ultralytics} to three dimensional radiological images and also uses some design choices of the more recent YOLO26~\cite{jocher2026ultralytics}. 
In the transition from 2D to 3D, we retained the core architecture and training scheme of YOLO11 while replacing 2D convolutions, batch normalization, and max pooling layers with their 3D counterparts.
\figurename~\ref{fig:radyolo_architecture_overview} shows an overview of the RadYOLO model architecture which consists of a feature extraction backbone, a feature pyramid neck and a detection and segmentation head.

We kept the anchor-free design and adapted the bounding box parametrization to predict the bounding box extent and center offset relative to its respective grid cell.
This allows RadYOLO to parametrize small objects more accurately than the original YOLO11 formulation, which predicts box face offsets relative to the grid cell.
We replaced the distribution focal loss (DFL) for bounding box prediction with an L1 loss, like it was done in YOLO26, and removed the respective DFL module to directly predict the box extent and center offset.

We added a small-object fallback to the task-aligned assigner (TAL), ensuring that small objects with zero overlap with any predicted box are still assigned to a box as a training target.

To allow simultaneous object detection and instance segmentation, we added a layer to predict prototype segmentation masks on half the input image resolution per axis and to predict mask coefficients for each prototype mask to the object detections.
The prototype masks are multiplied by the mask coefficients, and a sigmoid activation is applied to produce instance segmentations, following Bolya et al.~\cite{bolya2019yolact}.
To obtain full-resolution segmentation masks, we upsample the half-resolution mask logits via trilinear interpolation and then binarize them.

As resampling strategy, RadYOLO supports both resampling to a fixed image extent for whole-image prediction in one pass, as well as resampling to a fixed voxel size for patch-based training and inference to handle large imaging volumes.
For input normalization, we apply image-wise z-score normalization to MRI or mixed datasets and dataset-wise z-score normalization to CT datasets (using precomputed mean, standard deviation, and clipping thresholds).

For model selection, we use an exponential moving average of the mean average precision (mAP) computed at intersection-over-union (IoU) thresholds from 0.1 to 0.95.
In 2D image processing, IoU thresholds typically start at 0.5. However, for 3D imaging, we found that an IoU of 0.1 still reflects acceptable bounding box localization for small objects with ill-defined boundaries (like lesions).

For data augmentation, we employ transformations specific to the medical imaging domain.
Specifically, RadYOLO incorporates random mirroring, zoom, rotation, cutout, and intensity augmentation, where the latter is implemented using the batchgenerators library~\cite{isensee2020batchgenerators}.

The hyperparameters of RadYOLO were optimized using random hyperparameter search on datasets from the Medical Segmentation Decathlon~\cite{antonelli2022medical}.
Like YOLO11, RadYOLO offers multiple model sizes: nano\,(n), small\,(s), medium\,(m), large\,(l), and extra-large\,(x).
They define the channel count and depth of the backbone, neck, and head.
In this study we only evaluate the nano and small variants. 
Both share the same depth, but the small variant uses twice as many channels per layer as the nano variant.

\subsection{Experiments}
We compared RadYOLO with two other models: nnDetection~\cite{baumgartner2021nndetection} and nnU-Net~\cite{isensee2021nnu}. 
Both are self-configuring deep learning pipelines for medical image analysis. 
They use a dataset fingerprint together with heuristics to define the preprocessing, model architecture and hyperparameters for a given dataset. 
While nnDetection is a detection model that predicts bounding boxes with confidence scores, nnU-Net predicts full segmentation masks. 

We used the updated ResEnc M presets~\cite{isensee2024nnu} to train nnU-Net and selected the low resolution models for AMOS22 and VerSe while we used the full resolution model for Liver Lesions, LUNA16 and MAMA-MIA.
To derive bounding boxes with confidence scores from nnU-Net for detection metric calculation, we performed connected component analysis on its predicted segmentation masks and used each component's volume (in liters) as the confidence score.

RadYOLO resamples the input images to a fixed, dataset-specific image extent (number of voxels, see patch size in \tablename~\ref{tab:result_efficiency_metrics}) and performs a one-pass prediction for all datasets except for LUNA16. 
For LUNA16, RadYOLO resampled images to a fixed voxel size and a patch-based inference scheme was employed, analogous to that of nnU-Net and nnDetection. 
The voxel size was chosen as the same as nnU-Net and nnDection, namely $0.7\times0.7\times1.3 \; \textrm{mm}^3$. 
The other voxel sizes for nnU-Net and nnDetection were $0.8\times0.8\times1.0 \; \textrm{mm}^3$ (Liver Lesions), $0.7\times0.7\times1.2 \; \textrm{mm}^3$ (MAMA-MIA), $0.8\times0.8\times2.0 \; \textrm{mm}^3$ (AMOS22, nnDetection), $1.7\times1.7\times3.3 \; \textrm{mm}^3$ (AMOS22, nnU-Net), $1.0\times0.9\times1.0 \; \textrm{mm}^3$ (VerSe, nnDetection), and $1.9\times1.8\times2.0 \; \textrm{mm}^3$ (VerSe, nnU-Net).

While RadYOLO and nnU-Net use only one trained model for prediction, nnDetection uses an ensemble of four different model checkpoints.
In addition it employs test-time-augmentation by flipping the inputs once along each axis, yielding eight combinations.
In preliminary experiments, disabling nnDetection's ensembling and test-time augmentation drastically decreased its detection performance.
We therefore retained these settings despite their substantially increased inference time.

We used the mean average precision at an intersection-over-union threshold of 0.1 (mAP@0.1) and the mean over mAP@0.1 to mAP@0.95 in steps of 0.05 (mAP@0.1-0.95) as evaluation metrics. 
In addition, we report a FROC score, computed as the average sensitivity at false-positive rates of 1/8, 1/4, 1/2, 1, 2, 4, and 8 per image. Metrics are reported without confidence intervals or statistical testing.

The inference with its associated inference time measurements for all models was executed as batch processing. 
Thus, the inference time measurements include image loading, preprocessing, model inference, and writing results to disk. 
They do not include pipeline startup time or model loading time.
The time measurements were taken on a deep learning cluster with Nvidia A100 GPUs and Intel Xeon Gold 5320 CPUs. 
We did not compare model training times because they were trained on differing hardware and are therefore not comparable.

\section{Results}

\begin{table}[tb]
\caption{Detection and segmentation metrics on test data. mAP@.1, mAP@.1-.95 and FROC are calculated on bounding boxes with confidence scores. Dice and HD-95 are computed for segmentation predictions and show mean $\pm$ standard deviation over the test cases, where HD-95 is the Hausdorff-95 distance. Best results per dataset are bold.}
\label{tab:result_metrics}
\centering
\begin{tabular}{|c|l|c|c|c|c|c|}
\hline
Dataset & Model & mAP@.1 & mAP@.1-.95 & FROC & Dice & HD-95 [mm] \\
\hline
 AMOS22 & nnDetection & 0.93 & 0.39 & 0.32 & - & - \\
       & nnU-Net & \textbf{0.99} & \textbf{0.84} & 0.96 & $\textbf{0.89}\pm0.05$ & $\textbf{4.4}\pm3.8$ \\
       & RadYOLO\,(n) & \textbf{0.99} & 0.61 & \textbf{0.99} & $0.77\pm0.06$ & $9.0\pm3.4$ \\
       & RadYOLO\,(s) & \textbf{0.99} & 0.65 & \textbf{0.99} & $0.78\pm0.06$ & $8.2\pm3.4$ \\
\hline
 Liver Lesions & nnDetection & 0.67 & 0.35 & 0.52 & - & - \\
              & nnU-Net & 0.72 & \textbf{0.51} & 0.49 & $\textbf{0.75}\pm0.28$ & $38\pm51$ \\
              & RadYOLO\,(n) & 0.76 & 0.42 & 0.61 & $0.69\pm0.28$ & $27\pm36$ \\
              & RadYOLO\,(s) & \textbf{0.79} & 0.46 & \textbf{0.64} & $0.71\pm0.26$ & $\textbf{26}\pm36$ \\
\hline
 LUNA16 & nnDetection & \textbf{0.71} & 0.27 & \textbf{0.71} & - & - \\
        & nnU-Net & 0.52 & 0.26 & 0.54 & $0.34\pm0.26$ & $121\pm82$ \\
        & RadYOLO\,(n) & 0.67 & 0.33 & 0.66 & $0.46\pm0.31$ & $88\pm80$ \\
        & RadYOLO\,(s) & 0.69 & \textbf{0.36} & 0.69 & $\textbf{0.49}\pm0.31$ & $\textbf{87}\pm81$ \\
\hline
MAMA-MIA & nnDetection & 0.91 & 0.50 & 0.94 & - & - \\
         & nnU-Net & 0.85 & 0.50 & 0.92 & $\textbf{0.77}\pm0.20$ & $43\pm65$ \\
         & RadYOLO\,(n) & \textbf{0.96} & \textbf{0.59} & 0.96 & $0.75\pm0.22$ & $\textbf{12}\pm22$ \\
         & RadYOLO\,(s) & \textbf{0.96} & \textbf{0.59} & \textbf{0.97} & $0.75\pm0.21$ & $13\pm23$ \\
\hline
 VerSe & nnDetection & 0.89 & 0.52 & 0.72 & - & - \\
       & nnU-Net & \textbf{0.92} & \textbf{0.81} & 0.89 & $\textbf{0.83}\pm0.13$ & $\textbf{3.4}\pm4.6$ \\
      & RadYOLO\,(n) & \textbf{0.92} & 0.65 & 0.91 & $0.65\pm0.22$ & $4.8\pm5.7$ \\
      & RadYOLO\,(s) & 0.91 & 0.67 & \textbf{0.93} & $0.66\pm0.25$ & $4.7\pm6.0$ \\
\hline
\end{tabular}
\end{table}

\begin{figure}[tb]
\centering
\includegraphics[width=\textwidth]{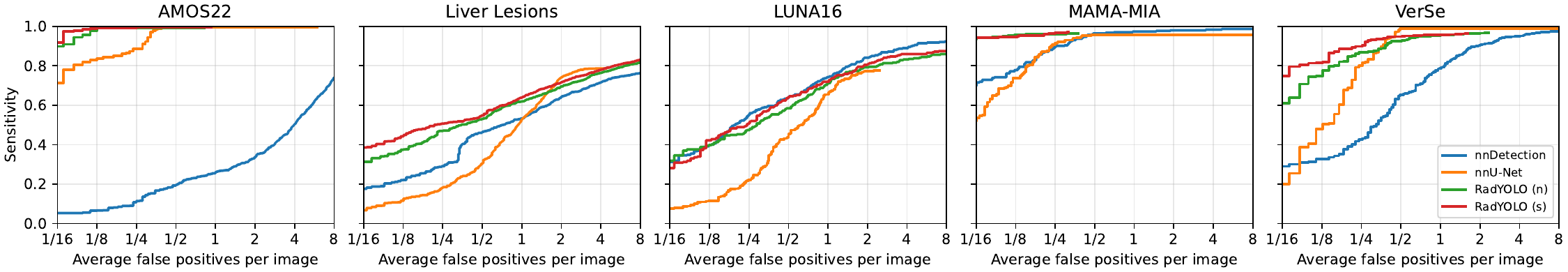}
\caption{FROC curves for IoU threshold of 0.1. nnU-Net reaches equal or higher sensitivity than RadYOLO only in the regime of 0.5–2 false positives or more per image.}
\label{fig:froc_curves}
\end{figure}

\tablename~\ref{tab:result_metrics} presents the detection and segmentation metrics. Regarding object detection, RadYOLO consistently outperforms nnDetection across all datasets except LUNA16, where there performance is on par for rough localization. 
Compared to nnU-Net, RadYOLO achieves higher or equal mAP@0.1 and FROC scores on all five datasets, indicating superior detection sensitivity when rough localization is sufficient.
\figurename~\ref{fig:froc_curves} shows the corresponding FROC curves.
On the lesion detection tasks (Liver Lesions, LUNA16, MAMA-MIA), RadYOLO outperforms nnU-Net across all detection metrics. 
For example, on LUNA16, RadYOLO (s) achieves a mAP@0.1 of 0.69 versus 0.52 for nnU-Net, and on MAMA-MIA 0.96 versus 0.85. 
On datasets with larger anatomical structures (AMOS22, VerSe), nnU-Net achieves substantially higher mAP@0.1-0.95 (0.84 vs. 0.65 on AMOS22; 0.81 vs. 0.67 on VerSe), reflecting its more precise bounding box localization for organs and vertebrae.

For segmentation, nnU-Net achieves higher Dice scores on AMOS22 (0.89 vs. 0.78) and VerSe (0.83 vs. 0.66). 
On lesion datasets, the gap is smaller (e.g., 0.75 vs. 0.71 on Liver Lesions), and RadYOLO achieves lower Hausdorff-95 distances on Liver Lesions (26 vs. 38\,mm) and MAMA-MIA (13 vs. 43\,mm), as well as higher Dice on LUNA16 (0.49 vs. 0.34). 
RadYOLO\,(s) generally performs slightly better than RadYOLO\,(n) across metrics.

\tablename~\ref{tab:result_efficiency_metrics} reports the efficiency metrics. 
On GPU, RadYOLO achieves speedups of 8–46$\times$ over nnU-Net and 36–236$\times$ over nnDetection depending on dataset. 
On CPU, RadYOLO processes cases in 1.0–33\,s, still outperforming the GPU runtimes of nnU-Net and nnDetection, while nnU-Net inference takes several minutes per case on CPU and nnDetection CPU inference times were prohibitively slow and thus not measured. 
RadYOLO uses 10\,M\,(n) or 38\,M\,(s) parameters compared to 192\,M for nnU-Net and 20\,M for nnDetection. 
Despite processing substantially larger patch sizes, RadYOLO's training GPU memory usage ranges from 5.1–42.7\,GB, comparable to the other models for most datasets. 
The models with high training memory requirements occur on datasets with many objects per image because RadYOLO performs instance segmentation.
During training, it predicts segmentation masks for each reference object for each assigned grid cell, which can accumulate to a large amount of memory if many reference objects are present.

\begin{table}[tb]
\caption{Computational efficiency metrics. Time GPU and Time CPU are the inference time to process a whole input volume using either GPU or CPU for model inference showing the mean $\pm$ standard deviation. Param. is the number of model parameters (M for million parameters). T.Mem. is the estimated (nnU-Net, nnDetection) or measured (RadYOLO) GPU memory spike during training. I.Mem. is the measured peak memory of one model inference pass using ONNX Runtime. MACs is the number of Multiply-Accumulate operations for one model inference pass (FLOPs = 2 $\times$ MACs) as estimated with onnx-tools.}
\label{tab:result_efficiency_metrics}
\scriptsize
\centering
\begin{tabular}{|c|l|c|c|r|r|r|r|c|c|}
\hline
Dataset & Model & Time GPU & Time CPU & Param. & T.Mem. & I.Mem. & MACs & Patch Size \\
\hline
 AMOS22 & nnDetection & $330 \pm 149$ s & - & 20\,M & 11\,GB & 1.5\,GB & 540\,G & $160\!\times\!160\!\times\!80$  \\
 & nnU-Net & $29\pm23$ s & $8.9\pm5.5$ m & 192\,M & 8\,GB & 1.6\,GB & 732\,G & $160\!\times\!160\!\times\!80$ \\
 & RadYOLO(n) & $1.8 \pm 1.2$ s & $2.0 \pm 1.4$ s & 10\,M & 13\,GB & 0.9\,GB & 77\,G & $352\!\times\!352\!\times\!128$  \\
 & RadYOLO(s) & $1.8 \pm 1.2$ s & $2.3 \pm 1.5$ s & 38\,M & 14\,GB & 1.0\,GB & 147\,G & $352\!\times\!352\!\times\!128$  \\
\hline
 LiverLesions & nnDetection & $65\pm25$ s & - & 19\,M & 11\,GB & 0.9\,GB & 320\,G & $160\!\times\!128\!\times\!96$  \\
 & nnU-Net & $23\pm12$ s & $22\pm12$ m & 192\,M & 8\,GB & 1.8\,GB & 878\,G & $160\!\times\!160\!\times\!96$  \\
 & RadYOLO(n) & $0.8 \pm 1.0$ s & $1.2 \pm 0.7$ s & 10\,M & 43\,GB & 0.8\,GB & 71\,G & $320\!\times\!288\!\times\!160$  \\
 & RadYOLO(s) & $0.5 \pm 0.6$ s & $1.8 \pm 1.0$ s & 38\,M & 25\,GB & 1.0\,GB & 137\,G & $320\!\times\!288\!\times\!160$  \\
\hline
 LUNA16 & nnDetection & $300\pm85$ s & - & 19\,M & 11\,GB & 1.0\,GB & 333\,G & $160\!\times\!160\!\times\!80$  \\
 & nnU-Net & $48\pm17$ s & $94\pm43$ m & 192\,M & 8\,GB & 1.8\,GB & 878\,G & $160\!\times\!192\!\times\!80$  \\
 & RadYOLO(n) & $6.3 \pm 6.3$ s & $24 \pm 14$ s & 10\,M & 10\,GB & 0.9\,GB & 78\,G & $320\!\times\!320\!\times\!160$  \\
 & RadYOLO(s) & $5.8 \pm 5.4$ s & $33 \pm 16$ s & 37\,M & 11\,GB & 1.1\,GB & 150\,G & $320\!\times\!320\!\times\!160$  \\
\hline
 MAMA-MIA & nnDetection & $118\pm85$ s & - & 19\,M & 11\,GB & 1.0\,GB & 335\,G & $160\!\times\!160\!\times\!80$  \\
 & nnU-Net & $10\pm11$ s & $18\pm18$ m & 192\,M & 8\,GB & 1.8\,GB & 880\,G & $160\!\times\!160\!\times\!96$  \\
 & RadYOLO(n) & $0.5 \pm 0.5$ s & $1.0 \pm 0.3$ s & 10\,M & 5\,GB & 0.6\,GB & 51\,G & $256\!\times\!256\!\times\!160$ \\
 & RadYOLO(s) & $0.4 \pm 0.5$ s & $1.2 \pm 0.3$ s & 38\,M & 6\,GB & 0.7\,GB & 98\,G & $256\!\times\!256\!\times\!160$  \\
\hline
 VerSe & nnDetection & $360\pm360$ s & - & 21\,M & 11\,GB & 1.0\,GB & 419\,G & $112\!\times\!112\!\times\!160$ \\
 & nnU-Net & $68\pm86$ s & $11 \pm 15$ m & 192\,M & 8\,GB & 1.7\,GB & 821\,G & $112\!\times\!128\!\times\!160$ \\
 & RadYOLO(n) & $4.3 \pm 4.9$ s & $4.8 \pm 4.9$ s & 10\,M & 19\,GB & 0.9\,GB & 81\,G & $256\!\times\!256\!\times\!256$ \\
 & RadYOLO(s) & $4.5 \pm 5.0$ s & $4.1 \pm 4.0$ s & 38\,M & 19\,GB & 1.1\,GB & 156\,G & $256\!\times\!256\!\times\!256$ \\
\hline
\end{tabular}
\end{table}

\section{Discussion}
RadYOLO demonstrates strong detection performance across diverse datasets while achieving substantial inference speedups. 
Its superiority over nnDetection is notable given that nnDetection relies on ensembling four model checkpoints with test-time augmentation, without which its performance degrades. 
This dependency makes nnDetection impractical for time-sensitive applications.

The detection results reveal a task-dependent trade-off between RadYOLO and nnU-Net. 
For lesion detection (Liver Lesions, LUNA16, MAMA-MIA), RadYOLO consistently outperforms nnU-Net across all detection metrics, likely because its explicit objectness prediction and confidence scoring are better suited to detect small, focal abnormalities than relying on post-hoc connected component analysis of segmentation masks.
More sophisticated post-processing might improve nnU-Net's detection results but would introduce additional complexity.
Conversely, for large anatomical structures (AMOS22, VerSe), nnU-Net achieves higher mAP@0.1-0.95, indicating more precise bounding box localization.  
Notably, at mAP@0.1, RadYOLO matches nnU-Net even on these datasets, confirming reliable detection when exact boundary delineation is less critical.

The segmentation gap on AMOS22 and VerSe can be attributed to RadYOLO's prototype mask approach, which predicts masks at half input resolution and relies on a limited set of learned prototypes.
This approach is inherently less expressive than nnU-Net's full-resolution, voxel-wise prediction.
However, on lesion datasets, RadYOLO achieves comparable or lower Hausdorff-95 distances (e.g., 13 vs. 43 mm on MAMA-MIA), suggesting that for compact objects with ill-defined boundaries, prototype-based segmentation is sufficient and may even produce fewer catastrophic outliers.

RadYOLO's efficiency gains are clinically relevant: CPU inference times of 1–33 seconds enable deployment without GPU hardware, which is critical for point-of-care or resource-constrained clinical environments.
The 5–19× reduction in parameters compared to nnU-Net further facilitates deployment on edge devices. 
However, training memory requirements remain a limitation for datasets with many objects per image due to the instance segmentation head.
This could be mitigated by omitting segmentation when only detection is needed.

\section{Conclusion}
We presented RadYOLO, a publicly available 3D extension of YOLO11 for joint object detection, classification, and segmentation in medical images.
Across five diverse datasets, RadYOLO matches or exceeds the detection performance of nnU-Net and consistently outperforms nnDetection, while being 8–46× faster on GPU and enabling CPU-only inference within seconds, even on large input volumes.
For lesion detection tasks, RadYOLO offers both superior detection sensitivity and competitive segmentation quality.
When precise localization of large organs is required, nnU-Net remains advantageous, suggesting the two approaches are complementary.
RadYOLO's combination of accuracy, speed, and broad applicability makes it a strong candidate for efficient clinical deployment of 3D medical image analysis.

\begin{credits}
\subsubsection{\ackname} This preprint has not undergone peer review or any post-submission improvements or corrections. The Version of Record of this contribution is published in [insert volume title when published], and is available online at https://doi.org/[insert DOI when published]

\subsubsection{\discintname}
The authors have no competing interests to declare that are
relevant to the content of this article.
\end{credits}

\bibliographystyle{splncs04}
\bibliography{references.bib}

\end{document}